\documentclass[journal,twoside]{IEEEtran}

\usepackage{cite}
\usepackage{amsmath,amssymb,amsfonts}
\usepackage{algorithmic}
\usepackage{graphicx}
\usepackage{textcomp}

\usepackage{multirow}
\usepackage{tabu}
\usepackage{url}
\usepackage{amssymb}
\usepackage{amsmath} 
\usepackage{cite}
\usepackage{float}
\usepackage{threeparttable}
\usepackage{color}

\begin{document}

\title{One-pass Multi-task Networks with Cross-task Guided Attention for Brain Tumor Segmentation}

\author{Chenhong Zhou,~\IEEEmembership{}
        Changxing Ding,~\IEEEmembership{}
        Xinchao Wang,~\IEEEmembership{}
        Zhentai Lu,~\IEEEmembership{}
        Dacheng Tao,~\IEEEmembership{}   
        
\thanks{Corresponding author: Changxing Ding.}  
\thanks{
 C. Zhou and C. Ding are with the School of Electronic and Information Engineering, South China University of Technology, Guangzhou, 510000, China (e-mail: eezhouch@mail.scut.edu.cn; chxding@scut.edu.cn).}
\thanks{X. Wang is with Department of Computer Science, Stevens Institute of Technology, Hoboken, USA.}
\thanks{Z. Lu is with Guangdong Provincial Key Laboratory of Medical Image Processing, School of Biomedical Engineering, Southern Medical University, Guangzhou, China.}
\thanks{D. Tao is with the UBTech Sydney Artificial Intelligence Centre and the School of Information Technologies, Faculty of Engineering and Information Technologies, The University of Sydney, Darlington, NSW 2008, Australia.}
}

\maketitle

\begin{abstract} 
Class imbalance has emerged as one of the major challenges for medical image segmentation. The model cascade (MC) strategy, a popular scheme, significantly alleviates the class imbalance issue via running a set of individual deep models for coarse-to-fine segmentation. Despite its outstanding performance, however, this method leads to undesired system complexity and also ignores the correlation among the models. To handle these flaws in the MC approach, we propose in this paper a light-weight deep model, i.e., the One-pass Multi-task Network (OM-Net) to solve class imbalance better than MC does, while requiring only one-pass computation for brain tumor segmentation. First, OM-Net integrates the separate segmentation tasks into one deep model, which consists of shared parameters to learn joint features, as well as task-specific parameters to learn discriminative features. Second, to more effectively optimize OM-Net, we take advantage of the correlation among tasks to design both an online training data transfer strategy and a curriculum learning-based training strategy. Third, we further propose sharing prediction results between tasks, which enables us to design a cross-task guided attention (CGA) module. By following the guidance of the prediction results provided by the previous task, CGA can adaptively recalibrate channel-wise feature responses based on the category-specific statistics. Finally, a simple yet effective post-processing method is introduced to refine the segmentation results of the proposed attention network. Extensive experiments are conducted to demonstrate the effectiveness of the proposed techniques. Most impressively, we achieve state-of-the-art performance on the BraTS 2015 testing set and BraTS 2017 online validation set. Using these proposed approaches, we also won joint third place in the BraTS 2018 challenge among 64 participating teams. The code is publicly available at
\url{https://github.com/chenhong-zhou/OM-Net}.

\end{abstract}


\IEEEpeerreviewmaketitle

\section{Introduction}

\IEEEPARstart{B}{rain} tumors are one of the most deadly cancers worldwide. Among these tumors, glioma is the most common type\cite{icsin2016review}. The average survival time for glioblastoma patients is less than 14 months \cite{van2010exciting}. Timely diagnosis of brain tumors is thus vital to ensuring appropriate treatment planning, surgery, and follow-up visits \cite{menze2015multimodal}. As a popular non-invasive technique, Magnetic Resonance Imaging (MRI)  produces markedly different types of tissue contrast and has thus been widely used by radiologists to diagnose brain tumors \cite{bauer2013survey}. However, the manual segmentation of brain tumors from MRI images is both subjective and time-consuming \cite{pereira2016brain}. Therefore, it is highly desirable to design automatic and robust brain tumor segmentation tools.

Recently, deep learning-based methods such as convolutional neural networks (CNNs) \cite{pereira2016brain,havaei2017brain,kamnitsas2017efficient,long2015fully,ronneberger2015u,chen2018voxresnet,zhao2018deep,kamnitsas2017ensembles,saha2018her2net,farag2017bottom,myronenko20183d,fehri2019bayesian,xiang2018automatic,zhou2018learning}, have become increasingly popular and achieved significant progress in brain tumor segmentation tasks. Unfortunately, a severe class imbalance problem usually emerges between healthy tissue and tumor tissue, as well as between intra-tumoral classes. This problem causes the healthy tissue to be dominant during the training phase and degrades the optimization quality of the model. To handle the class imbalance problem, many recent studies have employed the Model Cascade (MC) strategy 
\cite{haghighi2018automatic,hu2017detection,christ2016automatic,zhu20173d,lessmann2018automatic,zhang2017detecting,guo2017framework,tang2017scene,wang2017automatic}. More specifically, the MC strategy decomposes medical image segmentation into two or more tasks, each of which is achieved by an individual model. The most common MC framework for segmentation tasks \cite{haghighi2018automatic,hu2017detection,christ2016automatic,zhu20173d,lessmann2018automatic,zhang2017detecting,guo2017framework} incorporates two models, where the first one detects regions of interest (ROIs) via coarse segmentation, while the second conducts fine segmentation within the ROIs. Therefore, MC can effectively alleviate class imbalance via coarse-to-fine segmentation. In spite of its effectiveness, however, MC has several disadvantages. First, it usually requires multiple deep models to be trained, which substantially increases both the system complexity and the storage space consumption. Second, each model is trained separately using its own training data, which ignores the correlation between the deep models. Third, MC runs the deep models one by one, which leads to alternate GPU-CPU computations and a lack of online interactions between tasks.

\begin{figure} 
	\centering
	\includegraphics[height=5.5cm,width=8.5cm]{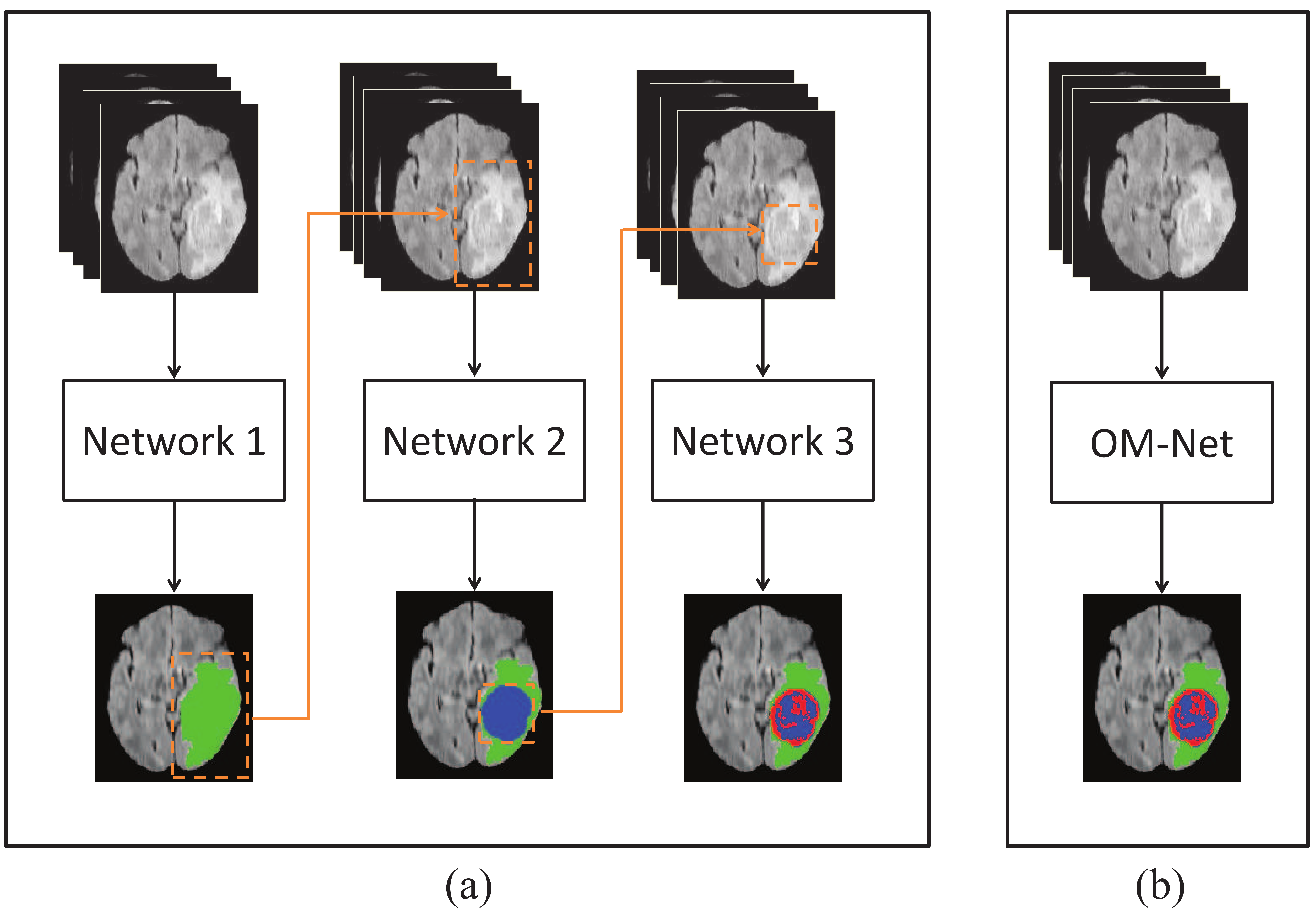}	
	\caption{Illustrations of (a) a three-model cascade pipeline and (b) our proposed OM-Net. The model cascade pipeline contains three networks that segment different tumor regions sequentially. OM-Net is a novel end-to-end deep model that simplifies prediction using one-pass computation. 
	}
	\label{Fig.0}
\end{figure}

Here, we propose adopting multi-task learning to overcome the shortcomings of MC. In more detail, we aim to decompose multi-class brain tumor segmentation into three separate yet interconnected tasks. While MC trains one individual network for each task, as shown in Fig. \ref{Fig.0}(a), we incorporate the three tasks into a single model and propose a One-pass Multi-task Network (OM-Net). The proposed OM-Net not only makes use of the relevance of these tasks to each other during the training stage, but also simplifies the prediction stage by implementing one-pass computation, as shown in Fig. \ref{Fig.0}(b). Furthermore, an effective training scheme inspired by curriculum learning is also designed: instead of training the three tasks together all the time, we gradually add the tasks to OM-Net in an increasing order of difficulty; this is beneficial for improving the convergence quality of the model.

Moreover, the fact that OM-Net integrates three tasks provides the possibility of online interaction between these tasks, which produces more benefits. First, the online training data transfer strategy we propose here enables the three tasks to share training data. Therefore, certain tasks obtain more training data, and the overall optimization quality can be improved. Second, we construct a novel channel attention module, named Cross-task Guided Attention (CGA), by sharing prediction results between tasks. In CGA, the prediction results of a preceding task can guide the following task to obtain category-specific statistics for each channel beforehand. This category-specific information further enables CGA to predict channel-wise dependencies with regard to a specific category of voxels. By contrast, existing self-attention models such as the popular `squeeze \& excitation' (SE) block \cite{Hu_2018_CVPR}, do not make use of such external guidance. Without external guidance, SE blocks only predict a single weight for each channel. However, there are usually multiple categories in a patch, and the importance of each channel varies for different categories. The proposed CGA module handles this problem by predicting the category-specific channel attention.

To further refine the segmentation results of OM-Net, we also propose a new post-processing scheme. The efficacy of the proposed methods is systematically evaluated on three popular brain tumor segmentation datasets, namely BraTS 2015, 2017, and 2018. Experimental results indicate that OM-Net outperforms MC, despite having only one-third of the model parameters of MC. The CGA module further promotes the performance of OM-Net by a significant margin.

A preliminary version of this paper has previously been published in \cite{zhou2018one}. Compared with the conference version, this version  proposes the novel CGA module, improves the post-processing method, and includes more experimental investigation. The remainder of this paper is organized as follows. We briefly review the related works for brain tumor segmentation in Section II, then provide the details of the OM-Net model in Section III. Experimental settings and datasets are detailed in Section IV, while the experimental results and analysis are presented in Section V. Finally, we conclude the paper in Section VI.

\section{Related Works}
In this section, we briefly review approaches in two domains related to our proposed model: namely, brain tumor segmentation and attention mechanism.

\subsection{Brain Tumor Segmentation} 
In recent years, deep learning-based methods such as CNNs have dominated the field of automatic brain tumor segmentation. The architectures of deep models \cite{long2015fully,ronneberger2015u, pereira2016brain, havaei2017brain,kamnitsas2017efficient, zhao2018deep, chen2018voxresnet, kamnitsas2017ensembles,saha2018her2net,farag2017bottom,myronenko20183d,fehri2019bayesian,xiang2018automatic,zhou2018learning} have developed rapidly from single-label prediction (classifying only the central voxel of the input patch) to dense-prediction (making predictions for all voxels in the input patch simultaneously). For instance, Pereira \emph{et al.} \cite{pereira2016brain} designed a deep model equipped with small convolutional kernels to classify the central voxel of the input 2D patch. Moreover, Havaei \emph{et al.} \cite{havaei2017brain} introduced a novel 2D two-pathway deep model to explore additional contextual information. The above methods make predictions based on 2D patches, ignoring the 3D contextual information. To handle this problem, Kamnitsas \emph{et al.} \cite{kamnitsas2017efficient}
introduced the DeepMedic model, which extracts information from 3D patches using 3D convolutional kernels. The abovementioned methods make predictions only for a single voxel or a set of central voxels within the input patch, meaning that they are slow in the inference stage. To promote efficiency, encoder-decoder architectures such as fully convolutional networks (FCNs) \cite{long2015fully} and U-Net \cite{ronneberger2015u} have been widely adopted to realize dense prediction. For instance, Chen \emph{et al.} \cite{chen2018voxresnet} designed a voxelwise residual network (VoxResNet) to make predictions for all voxels within the input 3D patch. Zhao \emph{et al.} \cite{zhao2018deep} introduced a unified framework integrating FCNs and conditional random fields (CRFs) \cite{krahenbuhl2011efficient}. This framework realizes end-to-end dense prediction with appearance and spatial consistency.

The issue of class imbalance is commonly encountered in medical image segmentation, especially brain tumor segmentation. To address this problem, many recent studies have adopted the MC strategy \cite{haghighi2018automatic,hu2017detection,christ2016automatic,zhu20173d,lessmann2018automatic,zhang2017detecting,guo2017framework,tang2017scene,wang2017automatic} to perform coarse-to-fine segmentation. In particular, the common two-model cascaded framework has been widely adopted in many applications, including renal segmentation in dynamic contrast-enhanced MRI (DCE-MRI) images \cite{haghighi2018automatic}, cancer cell detection in phase contrast microscopy images \cite{ hu2017detection }, liver and lesion segmentation \cite{ christ2016automatic }, volumetric pancreas segmentation in CT images \cite{ zhu20173d}, calcium scoring in low-dose chest CT images \cite{lessmann2018automatic}, etc. Moreover, MC can incorporate more stages to achieve better segmentation performance. For instance, Wang \emph{et al.} \cite{wang2017automatic} divided the brain tumor segmentation into three successive binary segmentation problems: namely, the segmentation of the complete tumor, tumor core, and enhancing tumor areas in MRI images. Since MC effectively alleviates class imbalance, its results are very encouraging.

Despite its effectiveness, however, MC is cumbersome in terms of system complexity; moreover, it ignores the correlation among tasks. Accordingly, in this paper, we adopt multi-task learning to overcome the disadvantages inherent in MC. By implementing the sharing of model parameters and training data, our proposed OM-Net outperforms MC using only one-third of the model parameters of MC.

\subsection{Attention Mechanism}
Attention is a popular tool in deep learning that highlights useful information in feature maps while suppressing irrelevant information. The majority of existing studies \cite{Hu_2018_CVPR,jaderberg2015spatial,li2018harmonious,wang2017residual,roy2018concurrent,pereira2018adaptive,zhu2019attention} belong to the category of self-attention models, meaning that they infer attentions based only on feature maps. These models can be roughly categorized into three types, namely hard regional attention, soft spatial attention, and channel attention. Moreover, some studies combine two or more types of attention in one unified model \cite{roy2018concurrent,wang2017residual,li2018harmonious}. The spatial transformer network (STN) \cite{jaderberg2015spatial} is a representative example of a hard attention model. STN selects and reshapes important regions in feature maps to a canonical pose to simplify inference. STN performs at the coarse region-level while neglecting the fine pixel-level saliency \cite{li2018harmonious}. In comparison, soft spatial attention models aim to evaluate pixel-wise importance in the spatial dimension. For instance, Wang \emph{et al.} \cite{wang2017residual} proposed a residual attention learning method that adds soft weights to feature maps via a residual unit, in order to refine the feature maps.

Complementary to spatial attention, channel attention aims to recalibrate channel-wise feature responses. The `squeeze \& excitation' (SE) block \cite{Hu_2018_CVPR} is one of the most popular channel attention models due to its simplicity and efficiency. However, this model was originally proposed for image classification and object detection tasks, but may not be optimal for image segmentation tasks; this is because SE blocks are based on the average response of all voxels in each channel and recalibrate each channel with a single weight, regardless of which category the voxels belong to. However, there are usually multiple categories of voxels in one patch, and the importance of each channel varies between the different categories. Several existing studies have already tried to alleviate the abovementioned problems experienced by SE blocks during segmentation tasks. For example, Pereira \emph{et al.} \cite{pereira2018adaptive} designed a segmentation SE (SegSE) block that produces a channel descriptor for each voxel in feature maps; consequently, the obtained channel attention map is of the same size as the feature maps. Voxel-wise multiplication between the feature maps and the attention map produces the re-weighted features.

Furthermore, the proposed CGA module also aims to solve the problems of SE blocks for segmentation. Unlike the existing self-attention models \cite{Hu_2018_CVPR,jaderberg2015spatial,li2018harmonious,wang2017residual,roy2018concurrent,pereira2018adaptive,zhu2019attention}, we make use of the special structure of OM-Net to provide cross-task guidance for the learning of category-specific channel attention.

\section{Method}
In this section, we first present a strong segmentation baseline based on MC, then introduce the model structure and training strategy of OM-Net. Next, we further explain the principles of OM-Net from the perspective of the attention mechanism, and subsequently propose the CGA module that promotes the performance of OM-Net by predicting robust channel attention. Finally, we propose a simple but effective post-processing method in order to refine the segmentation results of the attention network.

\begin{figure*}
	\centering
	\includegraphics[height=6cm,width=14cm]{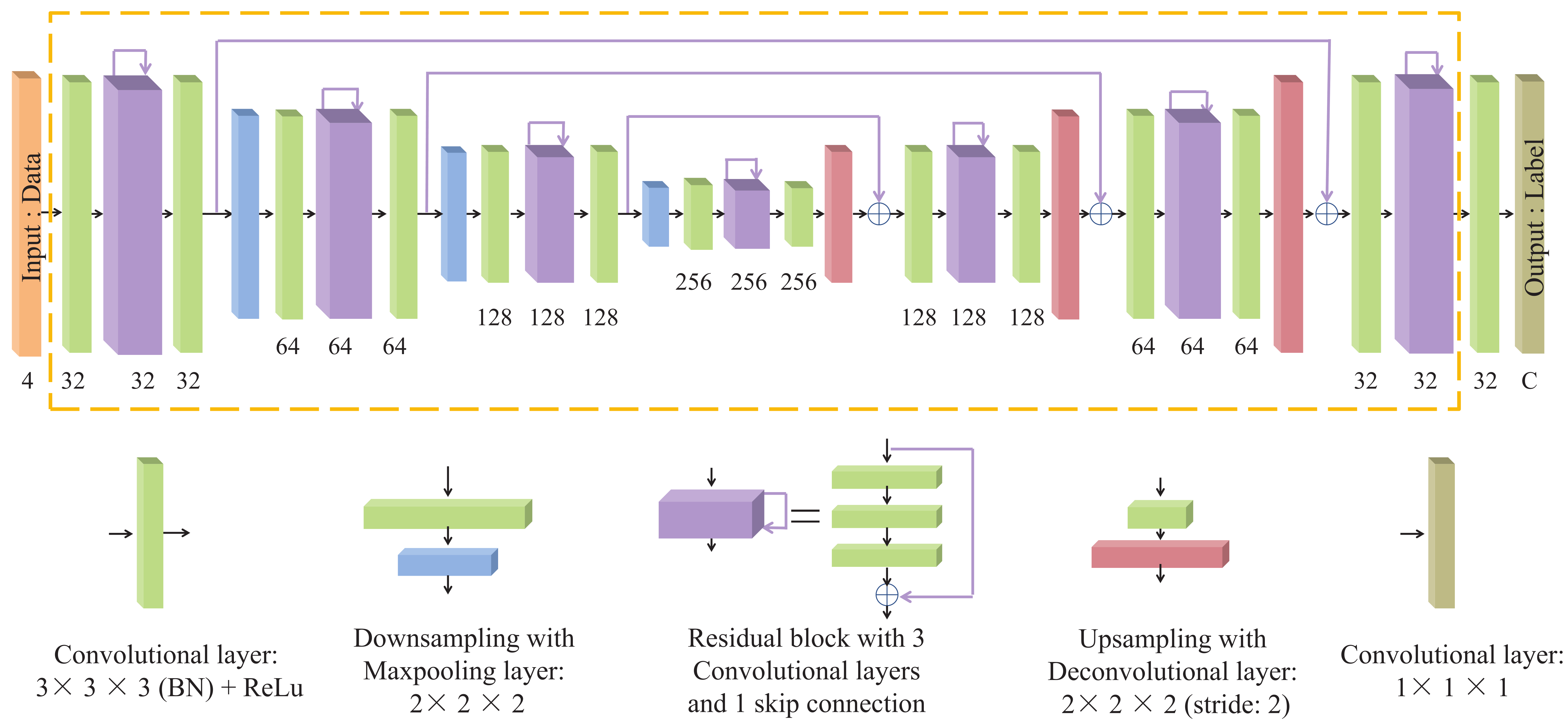}	
	\caption{The network structure for each task, which is composed of five basic building blocks. Each block is represented by a different type of colored cube. The number below each cube refers to the number of feature maps. C equals to 5, 5, and 2 for the first, second, and third task respectively. SoftmaxWithLoss is adopted as the loss function and applied to the output of each task. (Best viewed in color)
	}
	\label{Fig.1}
\end{figure*}

\subsection{A Strong Baseline Based on Model Cascade}

According to \cite{menze2015multimodal}, tumors can be divided into the following classes: edema (ED), necrotic (NCR), non-enhancing tumor (NET), and enhancing tumor (ET). Following \cite{brats2018_data}, we consistently merge NCR and NET into one class. Performance is evaluated on three re-defined tumor regions: complete tumor (including all tumor classes), tumor core (including all tumor classes except edema), and enhancing tumor (including only the enhancing tumor class). Under this definition, three regions satisfy the hierarchical structure of tumor subregions, each of which completely covers the subsequent one. Based on this observation, brain tumor segmentation can be decomposed into three separate yet interconnected tasks. In the following, we design an MC model that includes three independent networks and forms a strong baseline for OM-Net. Each network is trained for a specific task. These three tasks are detailed below.

1) \emph{Coarse segmentation of the complete tumor.} 
We utilize the first network to detect the complete tumor region as an ROI. We randomly sample training patches within the brain and train the network as a \emph{five-class segmentation task}: three tumor classes, normal tissue, and background. In the testing stage, we simply sum the predicted probabilities of all tumor classes to obtain a coarse tumor mask.
2) \emph{Refined segmentation for the complete tumor and its intra-tumoral classes}. We dilate the above coarse tumor mask by 5 voxels in order to reduce false negatives. Next, the labels of all voxels in the dilated region are predicted again \emph{as a five-class segmentation task} by the second network. Training data are sampled randomly within the dilated ground-truth complete tumor area. 
3) \emph{Precise segmentation for the enhancing tumor.} Due to the extreme class imbalance, it is very difficult to conduct precise segmentation of the enhancing tumor. To handle this problem, a third network is introduced specially for the segmentation of enhancing tumor. Similarly, the training patches for this task are  randomly sampled within the ground-truth tumor core area. It should be noted here that training patches for the three tasks are sampled independently, even if the sampling areas for the three tasks are nested hierarchically.

The network structures for the above three tasks are the same except for the final classification layer. The adopted structure is a 3D variant of the FusionNet \cite{quan2016fusionnet}, as shown in Fig. \ref{Fig.1}. We crop the MRI images to patches of size $32 \times 32 \times 16 \times 4$ voxels as input for the network. Here, the first three numbers correspond to the input volume, while the last number (4) denotes the four MRI modalities: FLAIR, T1-weighted (T1), T1 with gadolinium enhancing contrast (T1c), and T2-weighted (T2). Due to the lack of contextual information, the segmentation results of the boundary voxels in the patch may be inaccurate. Accordingly, we adopt the overlap-tile strategy proposed in \cite{ronneberger2015u} during inference. In brief, we sample 3D patches with a stride of $20 \times 20 \times 5$ voxels within the MRI image. For each patch, we retain only the predictions of voxels within the central region ($20 \times 20 \times 5$ voxels) and abandon the predictions of the boundary voxels. The predictions of all central regions from the sampled patches are stitched together to constitute the segmentation results of the whole brain. This strategy is also utilized in the following models.

During the inference stage of MC, the three networks have to be run one by one since the ROI of one task is obtained by considering the results of all preceding tasks. More specifically, we employ the first network to generate a coarse mask for the complete tumor; subsequently, all voxels within the dilated area of the mask are classified by the second network, from which the precise complete tumor and tumor core areas can be obtained. Finally, the third network is utilized to scan all voxels in the tumor core region in order to determine the precise enhancing tumor area. Thus, there are three alternate GPU-CPU computations carried out during the MC inference process.

\subsection{One-pass Multi-task Network}
Despite its promising performance, MC not only encounters disadvantages due to its system complexity, but also neglects the relevance among tasks. We can observe that the essential difference among these tasks lies in the training data rather than the model architecture. Therefore, we propose a multi-task learning model that integrates the three tasks involved in MC into one network. Each task in this model has its own training data that is exactly the same as the data in MC. Moreover, each task has an independent convolutional layer, a classification layer, and a loss layer. The other parameters are shared to make use of the correlation among the tasks. Thanks to the multi-task model, we can obtain the prediction results of the three classifiers simultaneously through one-pass computation. As a result, we name the proposed model the One-pass Multi-task Network, or OM-Net.

\begin{figure*}
	\centering
	\includegraphics[height=5.5cm,width=15cm]{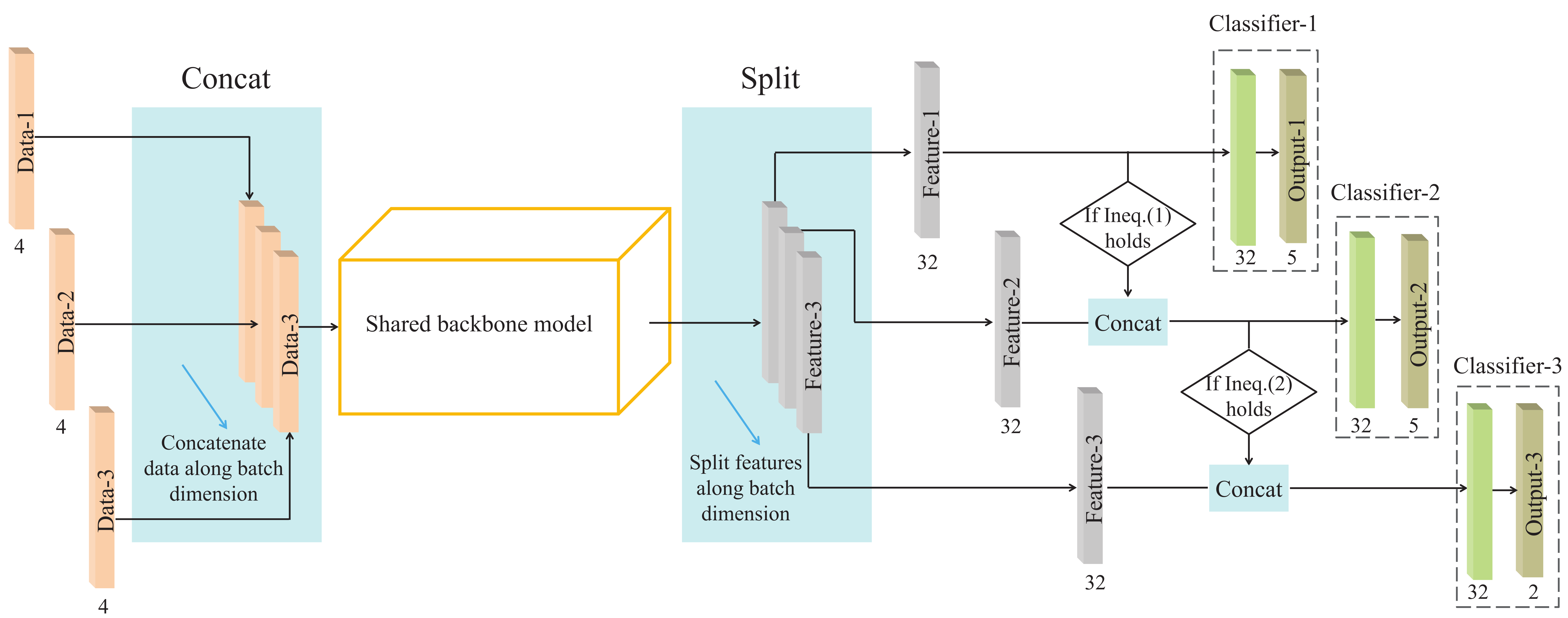}  
	\caption{Network structure of OM-Net in the training stage. For the i-th task, its training data, feature maps, and outputs of the classification layer are denoted as Data-$i$, Feature-$i$, and Output-$i$, respectively. The light blue rectangles marked with `Concat' and `Split' represent the concatenation and splitting operations, respectively, while the blue arrows represent the batch dimension.
		SoftmaxWithLoss is adopted as the loss function and applied to the output of each task. The shared backbone model refers to the network layers outlined by the yellow dashed line in Fig. \ref{Fig.1}. 
	}
	\label{Fig.2}
\end{figure*}

As the difficulty levels of these three tasks progressively increase, we propose to train OM-Net more effectively by employing curriculum learning\cite{bengio2009curriculum}, which is useful for improving the convergence quality of machine learning models. More specifically, instead of training the three tasks together all the time, we gradually introduce the tasks to the model in an order of increasing difficulty. The model structure and training strategy of OM-Net are illustrated in Fig. \ref{Fig.2}. First, OM-Net is trained with only the first task in order to learn the basic knowledge required to differentiate between tumors and normal tissue. This training process lasts until the loss curve displays a flattening trend.

The second task is then added to OM-Net; this means that the first and the second tasks are trained together. As illustrated in Fig. \ref{Fig.2}, we concatenate Data-1 and Data-2 along the batch dimension to form the input for OM-Net. We split the features generated by the shared backbone model along the batch dimension; here, the splitting position in the batch dimension is the same as the concatenation position of training data. We then obtain task-specific features and use the sliced features to optimize  task-specific parameters. Moreover, we argue that not only knowledge (model parameters) but also learning material (training data) can be transferred from the easier course (task) to the more difficult course (task) during curriculum learning. Since the sampling areas for the three tasks are hierarchically nested, we propose the following online training data transfer strategy. The training patches in Data-1 that satisfy the following sampling condition can be transferred to assist the training of the second task:
\begin{equation}
\frac{\sum\limits_{i=1}^{N}\mathbf{1}\left \{ l_i\in C_{complete} \right \}}{N} \geq 0.4 ,  \label{eq1}\
\end{equation}    
where $l_i$ is the label of the $i$-th voxel in the patch, $C_{complete}$ denotes the set of all tumor classes, $N$ is the number of voxels in the input patch, and $0.4$ is set to meet the patch sampling condition of the second task. We thus concatenate the features of these patches in Data-1 with Feature-2, then compute the loss for the second task. The training process in this step continues until the loss curve of the second task displays a flattening trend.

Finally, the third task and its training data are introduced to OM-Net, meaning that these three tasks are trained together. The concatenation and slicing operations are similar to those in the second step. The training patches from Data-1 and Data-2 that satisfy the following sampling condition can be transferred to the third task:
\begin{equation}
\frac{\sum\limits_{i=1}^{N}\mathbf{1}\left \{ l_i\in C_{core} \right \}}{N} \geq 0.5 ,   \label{eq2}\
\end{equation}   
where $C_{core}$ indicates the tumor classes belonging to the tumor core. Similarly, $0.5$ is chosen to meet the patch sampling condition of the third task. The threshold in Ineq. (1) is smaller than that in Ineq. (2); this is because the center points of training patches for the second task are sampled within the dilated area of the complete tumor. By comparison, the center points of training patches for the third task are sampled within the ground-truth area of the tumor core. This means that one training patch for the second task may include more than 50\% non-tumor voxels, while one training patch for the third task includes at most 50\% non-core voxels. Therefore, we set the thresholds in Ineq. (1) and Ineq. (2) to 0.4 and 0.5, respectively. The three tasks are trained together until convergence occurs. In conclusion, the OM-Net equipped with the curriculum learning-based training strategy has three main components: 1) a deep model based on multi-task learning that realizes coarse-to-fine segmentation via one-pass computation; 2) a stepwise training scheme that progresses from easy to difficult; 3) the transfer of training data from the easier task to the more difficult task.

In the inference stage, the data concatenation, feature slicing, and data transfer operations in Fig. \ref{Fig.2} are removed, and the 3D patches of an MRI image are fed into the shared backbone model. Feature-1, Feature-2, and Feature-3 are now the same for each patch. The prediction results of the three tasks can be obtained simultaneously by OM-Net. These results are fused in exactly the same way as in the MC baseline. Moreover, OM-Net is different from the existing multi-task learning models for brain tumor segmentation \cite{shen2017multi,shen2017boundary}. 
The principle behind these models \cite{shen2017multi,shen2017boundary} involves the provision of multiple supervisions for the same training data. By contrast, OM-Net aims to achieve coarse-to-fine segmentation by
integrating tasks with their own training data into a single model.

\subsection{Cross-task Guided Attention}

The coarse-to-fine segmentation strategy adopted by OM-Net can be regarded as a type of cascaded spatial attention, since the segmentation results of one task determine the ROI for the following task. In the following, we further enhance the performance of OM-Net from the perspective of channel attention. In particular, we propose a novel and effective channel attention model that makes use of cross-task guidance to solve the problems experienced by the popular SE block for the segmentation task.

\begin{figure}
	\centering 
	\includegraphics[width=\columnwidth]{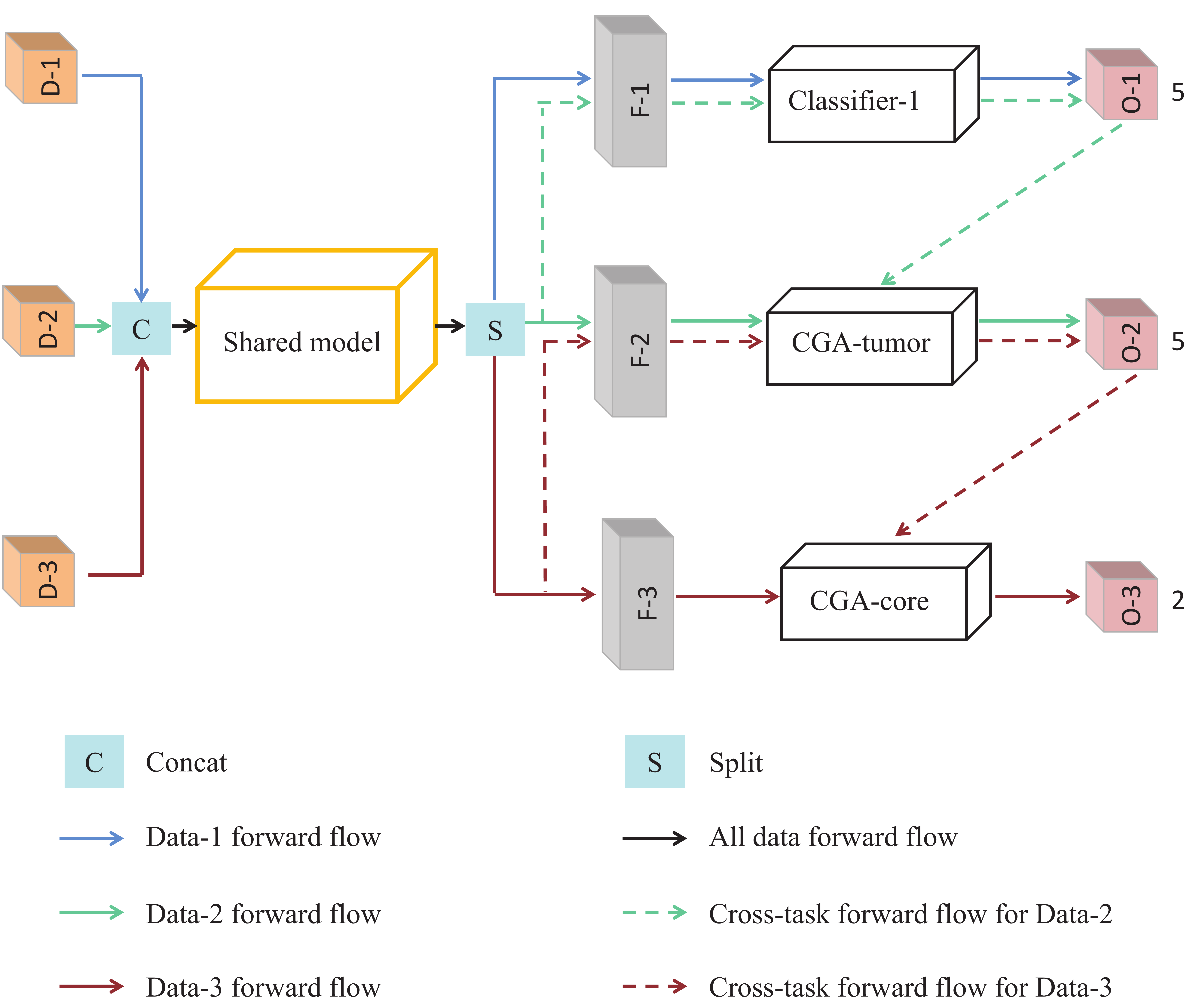}
	\caption{The overall architecture of OM-Net equipped with CGA modules in the training stage. D-$i$, F-$i$, and O-$i$ are abbreviations for Data-$i$, Feature-$i$, and Output-$i$, which denote the training data, features, and outputs for the $i$-th task, respectively. Feature-2 and Feature-3 pass through the first and second task respectively to obtain their own coarse prediction results beforehand. We refer to these two operations as cross-task forward flow for Data-2 and Data-3, represented by the green and red dotted lines respectively. These coarse predictions are utilized as cross-task guidance to help generate category-specific channel attention. SoftmaxWithLoss is adopted as the loss function and applied to the output of each task. The training data transfer strategy described in Fig. \ref{Fig.2} is omitted for clarity in this figure. 
	}
	\label{Fig.3}
	
\end{figure}

As explained in Section II, the global average pooling (GAP) operation in the SE block ignores the dramatic variation in volume of each class within the input patch. We solve this problem by computing statistics in category-specific regions rather than in a whole patch. However, the category-specific regions for common CNNs are unknown until we reach the final classification layer; therefore, this is a chicken-and-egg problem. Fortunately, OM-Net allows us to estimate category-specific regions beforehand by sharing the prediction results between tasks. More specifically, in the training stage, we let Feature-2 and Feature-3 (shown in Fig. \ref{Fig.2}) pass through the first and second task of OM-Net, respectively. In this way, we obtain the coarse segmentation results for the second and third tasks respectively. It is worth noting that this strategy introduces only negligible additional computation in the training stage and no extra computation in the testing stage. This is because Feature-1, Feature-2, and Feature-3 are exactly the same in the testing stage, as the concatenation and slicing operations in Fig. \ref{Fig.2} are removed during testing. 
Since we introduce cross-task guidance for the proposed channel attention block, we refer to it as Cross-task Guided Attention (CGA). We also rename Classifier-2 and Classifier-3 in Fig. \ref{Fig.2} once equipped with CGA as CGA-tumor and CGA-core, respectively. The overall architecture of OM-Net equipped with CGA modules is illustrated in Fig. \ref{Fig.3}. Note that cross-task guidance takes place only in the forward pass, meaning that the back-propagations of the three tasks are still independent. 

\begin{figure}
	\centering 
	\includegraphics[width=\columnwidth]{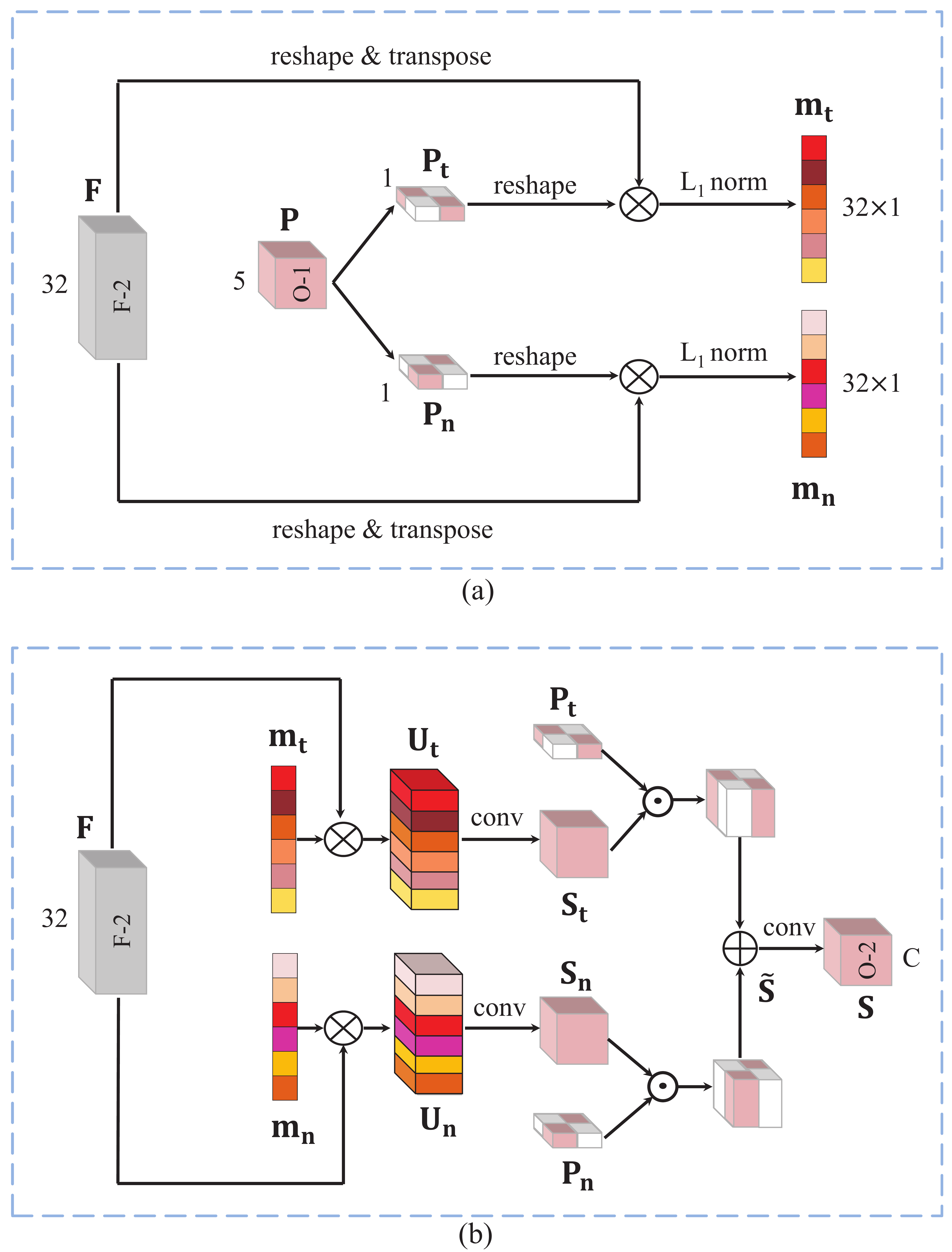} 	
	\caption{Structure of the CGA-tumor module. (a) Category-specific channel importance (CSCI) block; (b) complementary segmentation (CompSeg) block. The elements in	$\mathbf{{m}_{t}}$ and $\mathbf{{m}_{n}}$ describe the importance of each channel for one category of voxels within the patch, which are utilized to recalibrate the channel-wise feature responses.
	}
	\label{Fig.4}	
\end{figure}

As illustrated in Fig. \ref{Fig.4}, we take CGA-tumor as an example to illustrate the structure of the CGA module. This module is composed of two blocks: (a) a category-specific channel importance (CSCI) block, and (b) a complementary segmentation (CompSeg) block. In Fig. \ref{Fig.4}, all 4D feature maps and probability maps are simplified into 3D cubes to enable better visualization. In more detail, the height of one cube denotes the number of channels in feature maps. $\mathbf{P} \in \mathbb{R}^{W \times H \times L \times 5}$ denotes a probability tensor predicted by the preceding task and is represented as a pink cube. The number 5 next to the cube for $\mathbf{P}$ refers to the number of classes, which we have explained in the MC baseline. The grey cube $\mathbf{F} \in \mathbb{R}^{W \times H \times L \times 32}$ denotes the input feature maps of the current task for the CGA module. Moreover, the number 32 next to the cube for $\mathbf{F}$ denotes its number of channels.

\subsubsection{CSCI Block}
As illustrated in Fig. \ref{Fig.4}(a), the CSCI block utilizes both $\mathbf{P}$ estimated by the preceding task and $\mathbf{F}$ to estimate the importance of each channel for the segmentation of tumor and non-tumor categories, respectively. More specifically, we first compute $\mathbf{P_t}$ and $\mathbf{P_n}$; each of these values refers to the probability of one voxel belonging to the tumor and non-tumor categories, respectively: 
\begin{equation}
\mathbf{P_t}(i,j,k) = \sum_{c\in C_{tumor}}\mathbf{P}(i,j,k,c) , \label{eq3}\
\end{equation}  
\begin{equation}
\mathbf{P_n}(i,j,k) = \sum_{c\in C_{non-tumor}}\mathbf{P}(i,j,k,c) , \label{eq4}\
\end{equation}
where $\left\{ \mathbf{P_t}, \mathbf{P_n}\right\} \in \mathbb{R}^{W \times H \times L \times 1}$. $C_{tumor}$ and $C_{non-tumor}$ refer to the sets of classes that belong to the tumor and non-tumor categories, respectively. $C_{tumor}$ includes all tumor classes, while $C_{non-tumor}$ contains normal tissue and background.
We then reshape $\mathbf{P_t}$ and $\mathbf{P_n}$ to $\mathbb{R}^{N \times 1}$, respectively, where $N$ is equal to $W \times H \times L$ and denotes the number of voxels in a patch. Similarly, $\mathbf{F}$ is reshaped to $\mathbb{R}^{N \times 32}$. Subsequently, we perform matrix multiplication between the reshaped $\mathbf{F}$ and the reshaped $\mathbf{P_t}$, then apply $L_1$ normalization for the obtained vector:   
\begin{equation}
\mathbf{{m}_{t}}(i)=\frac{{Re(\mathbf{F})^T}_{i,:}\cdot{{Re(\mathbf{P_t}})}}{\sum\limits_{k=1}^{32}{Re(\mathbf{F})^T}_{k,:}\cdot{{Re(\mathbf{P_t}})}} , \label{eq5}\ 
\end{equation}
where $\mathbf{{m}_{t}} \in \mathbb{R}^{32 \times 1}$, while $Re(\cdot)$ denotes the reshape operation.
Similarly,
\begin{equation}
\mathbf{{m}_{n}}(i)=\frac{{Re(\mathbf{F})^T}_{i,:}\cdot{{Re(\mathbf{P_n}})}}{\sum\limits_{k=1}^{32}{Re(\mathbf{F})^T}_{k,:}\cdot{{Re(\mathbf{P_n}})}} . \label{eq5_1}\ 
\end{equation}
Elements in $\mathbf{{m}_{t}}$ and $\mathbf{{m}_{n}}$ describe the importance of each channel for the segmentation of tumor and non-tumor categories, respectively. Compared with the popular SE block, which squeezes the global information of each channel into a single value in order to describe its importance, CGA makes use of finer category-specific statistics to evaluate the importance of each channel for one specific category.

\subsubsection{CompSeg Block} 
Inspired by \cite {dey2018compnet}, we further propose a complementary segmentation (CompSeg) block that performs segmentation via two complementary pathways. These two pathways can make full use of the category-specific channel importance information ($\mathbf{{m}_{t}}$ and $\mathbf{{m}_{n}}$) to improve segmentation performance. As shown in Fig. \ref{Fig.4}(b), the two pathways focus on the segmentation of the tumor and non-tumor voxels respectively. The CompSeg block can be described in more detail as follows. First, $\mathbf{{m}_{t}}$ and $\mathbf{{m}_{n}}$ are used to recalibrate each channel in $\mathbf{F}$, respectively: 
\begin{equation}
\mathbf{U_t} = \mathcal{F}_{scale}(\mathbf{{m}_{t}}, \mathbf{F}) = \left[m_{t}^{1}\mathbf{f}_{1}, m_{t}^2\mathbf{f}_{2}, \cdot\cdot\cdot, m_{t}^{32}\mathbf{f}_{32}\right], \label{eq6}\
\end{equation}
\begin{equation}
\mathbf{U_n} = \mathcal{F}_{scale}(\mathbf{{m}_{n}}, \mathbf{F}) = \left[m_{n}^{1}\mathbf{f}_{1}, m_{n}^2\mathbf{f}_{2}, \cdot\cdot\cdot, m_{n}^{32}\mathbf{f}_{32}\right], \label{eq6_1}\
\end{equation}
where $\mathbf{f}_{i} \in \mathbb{R}^{W \times H \times L}$ is the $i$-th channel in $\mathbf{F}$. $m_{t}^{i}$ and $m_{n}^{i}$ are the $i$-th elements in $\mathbf{{m}_{t}}$ and $\mathbf{{m}_{n}}$, respectively. The recalibrated feature maps $\mathbf{U_t}$ and $\mathbf{U_n}$ highlight the more important channels and suppress the less important ones for tumors and non-tumors, respectively. These maps are then individually fed into a $1 \times 1 \times 1$ convolutional classification layer to produce their own score maps, $\mathbf{S_t}$ and $\mathbf{S_n}$. $\left\{ \mathbf{S_t}, \mathbf{S_n}\right\} \in \mathbb{R}^{W \times H \times L \times C}$, where $C$ refers to the number of classes for the current task. The two score maps are more sensitive to tumor and non-tumor classes respectively. Therefore, we merge the two score maps via weighted averaging:
\begin{equation}
\tilde{\mathbf{S}}(i,j,k,c) = {\mathbf{P_t}}(i,j,k) \cdot {\mathbf{S_t}}(i,j,k,c) + {\mathbf{P_n}}(i,j,k) \cdot {\mathbf{S_n}}(i,j,k,c) , \label{eq7}\
\end{equation}
where $\tilde{\mathbf{S}} \in \mathbb{R}^{W \times H \times L \times C}$. Finally, we feed $\tilde{\mathbf{S}}$ into another  $1 \times 1 \times 1$ convolutional layer to obtain the ultimate prediction results $\mathbf{S}$ for the current task.

As illustrated in Fig. \ref{Fig.4}, $\mathbf{P_t}$ and $\mathbf{P_n}$ are used twice in the CGA module. 
The first time, we use $\mathbf{P_t}$ and $\mathbf{P_n}$ in the CSCI block to provide category-specific probabilities, by which we calculate $\mathbf{{m}_{t}}$ and $\mathbf{{m}_{n}}$; these two vectors embed the interdependencies between channels with regard to different categories. The second time, we use them in the CompSeg block as soft spatial masks to merge two score maps by means of weighted averaging and produce the final segmentation results. Because all tasks are integrated in OM-Net, we are able to obtain $\mathbf{P_t}$ and $\mathbf{P_n}$ for use as cross-task guidance to compute category-specific statistics for each channel, thereby obtaining improved channel attentions. By comparison, the popular SE block ignores category-specific statistics and reweights each channel with a single weight. In the experiment section, we justify the effectiveness of both the CSCI block and the CompSeg block.

The model structures of the CGA-tumor and CGA-core modules are almost the same; there are only two trivial and intuitive differences. As illustrated in Fig. \ref{Fig.3}, the first difference lies in the position where we introduce the cross-task guidance; for the CGA-core module, it is introduced from the second task of OM-Net. Second, the counterparts of $\mathbf{P_t}$ and $\mathbf{P_n}$ in CGA-core indicate the probability of each voxel belonging to the core and non-core tumor categories, respectively.

\subsection{Post-processing}
We can observe from many prior studies \cite{pereira2016brain,kamnitsas2017efficient,havaei2017brain,zhao2018deep,kamnitsas2017ensembles,isensee2018no,chen2018focus} that post-processing is an efficient way to improve segmentation performance by refining the results of CNNs. For example, some small clusters of the predicted tumors are removed in \cite{pereira2016brain,kamnitsas2017ensembles,isensee2018no,chen2018focus}. In addition, conditional random field (CRF) is commonly used as a post-processing step in \cite{kamnitsas2017efficient,havaei2017brain}. In particular, Zhao \emph{et al.} \cite{zhao2018deep} proposed a post-processing method comprising six steps to boost the segmentation performance by a large margin.

In this paper, we introduce a simple and flexible post-processing method in order to refine the predictions of the proposed networks. Our method is mainly inspired by \cite{zhao2018deep}, but consists of fewer steps and adopts K-means clustering to achieve automatic classification rather than defining the thresholds of voxel intensities in \cite{zhao2018deep}.

Step 1:  We remove isolated small clusters with volumes smaller than a threshold $\tau_{VOL}$. ${\tau}_{VOL} = min (2000, 0.1 \times V_{max})$, where $V_{max}$ denotes the volume of the largest 3D connected tumor area predicted by the proposed model. This step can slightly improve the Dice score for the complete tumor, as false positives are removed.

Step 2: It is observed that non-enhancing voxels are likely to be misclassified as edema if the predicted enhancing tumor area is small \cite{zhao2018deep}. Accordingly, we propose a K-means-based method to handle this problem, as follows.

Let $vol_e$ and $vol_t$ denote the volumes of enhancing tumor and complete tumor in the predicted results, respectively. Moreover, $vol_e(n)$ and $vol_t(n)$ refer to the volumes of enhancing tumor and complete tumor in the $n$-th 3D connected tumor area, respectively. If $vol_e/vol_t < 0.1$, $vol_e(n)/vol_t(n) < 0.05$, and $vol_e(n) < 1000$, the K-means clustering algorithm is employed. Based on their intensity values in the MRI images, the segmented edema voxels in the $n$-th connected component are clustered into two groups. Finally, the average intensity of each group in the T1c channel is computed. We convert the labels of voxels in the group with the lower averaged intensity to the non-enhancing class. The labels of the voxels in the other group remain unchanged.

In the experiment section, we show that the second step significantly improves the Dice score of the tumor core. As this step only changes the labels of the voxels predicted as edema, it will not affect the segmentation results of the complete tumor or enhancing tumor.

\section{Experimental Setup}
In this section, we provide the datasets used to validate our approaches, the evaluation metrics, and the implementation details.

\subsection{Datasets}
To demonstrate the effectiveness of the proposed methods, we conduct experiments on the BraTS 2018 \cite{menze2015multimodal, bakas2017advancing, bakas2017segmentationlgg, bakas2017segmentationgbm,bakas2018identifying}, BraTS 2017 \cite{menze2015multimodal, bakas2017advancing, bakas2017segmentationlgg, bakas2017segmentationgbm}, and BraTS 2015 \cite{menze2015multimodal, kistler2013virtual} datasets. There are four modalities for each MRI image. All images in the three datasets have been co-registered, interpolated and skull-stripped. The dimensions of all images are $240 \times 240 \times 155$ voxels.

The BraTS 2018 dataset contains three subsets: the training set, testing set, and validation set. The training set comprises 210 cases of high-grade gliomas (HGG) and 75 cases of low-grade gliomas (LGG). The testing and validation sets contain 191 cases and 66 cases respectively with hidden ground-truth. The evaluation metrics of the testing and validation sets are computed using an online evaluation platform \cite{CBICAIPP2015}.

The BraTS 2017 dataset shares an identical training set with BraTS 2018. Compared with BraTS 2018, it has a smaller validation set comprising 46 cases. The evaluation of the validation set is conducted online \cite{CBICAIPP2015}.

The BraTS 2015 dataset consists of a training set including 274 MRI images and a testing set including 110 MRI images. Performance evaluation of the testing set is also conducted using an online evaluation platform \cite{VirtualSkeleton2013}.

\subsection{Evaluation Metrics}
We follow the official evaluation metrics for each dataset. There are a number of different metrics, namely the Dice score, Positive Predictive Value (PPV), Sensitivity, and Hausdorff distance, each of which is defined below:
\begin{equation}
Dice = \frac{2TP}{FP + 2TP + FN} , \label{eq8}\
\end{equation}  
\begin{equation}
PPV = \frac{TP}{FP + TP} , \label{eq9}\
\end{equation}  
\begin{equation}
Sensitivity = \frac{TP}{TP + FN} , \label{eq10}\
\end{equation}   
\begin{equation}
Haus(T,P) = max\small\{\mathop{sup }\limits_{t\in {T} } \mathop{inf}\limits_{p\in {P}} d(t,p), \mathop{sup}\limits_{p\in {P} } \mathop{inf}\limits_{t\in {T}} d(t,p)\small\} , \label{eq11}\
\end{equation} 
where the number of false negative, true negative, true positive, and false positive voxels are denoted as FN, TN, TP, and FP, respectively. \emph{sup} represents the supremum and \emph{inf} denotes the infimum, while $t$ and $p$ denote the points on surface ${T}$ of the ground-truth regions and surface ${P}$ of the predicted regions, respectively. $d(\cdot,\cdot)$ is the function that computes the distance between points $t$ and $p$. 
Dice score, PPV, and Sensitivity measure the voxel-wise overlap between the ground-truth and the predicted results \cite{menze2015multimodal}. The Hausdorff distance evaluates the  distance between the surface of the ground-truth regions and that of the predicted regions. Moreover, Hausdorff95 is a metric of Hausdorff distance used to measure the 95\% quantile of the surface distance. As the Dice score is the overall evaluation metric, adopted consistently across all the BraTS challenges, we adopt it as the main metric for evaluation in line with existing works \cite{pereira2016brain,havaei2017brain,zhao2018deep,kamnitsas2017efficient,kamnitsas2017ensembles,myronenko20183d,wang2017automatic,pereira2018adaptive,shen2017multi,shen2017boundary,bakas2018identifying,pereira2018adaptive,chen2018focus,isensee2018no}.

\subsection{Implementation Details}

During pre-processing, we normalize the voxel intensities within the brain area to have zero mean and unit variance for each MRI modality. The numbers of training patches are around 400,000, 400,000, and 200,000 for the first, second, and third task respectively. SoftmaxWithLoss is adopted as the loss function consistently. All implementations are based on the \emph{C3D}\footnote[1]{\url{https://github.com/facebook/C3D}} \cite{tran2015learning,karpathy2014large,jia2014caffe} package, which is a modified 3D version of \emph{Caffe}\cite{jia2014caffe}. The models are trained using stochastic gradient descent with a momentum of 0.99 and a batchsize of 20 for each task. The learning rate of all networks is initially set to 0.001, which is divided by 2 after every four epochs. We train each network in MC for 20 epochs. Similarly, we train OM-Net for 1 epoch, 1 epoch, and 18 epochs for its three steps, respectively. Therefore, the three tasks in OM-Net are trained for 20, 19, and 18 epochs, respectively.

\section{Experimental Results and Discussion}
We first carry out ablation studies to demonstrate the validity of each contribution proposed in this paper. We then compare the performance of the proposed methods with state-of-the-art brain tumor segmentation approaches on the BraTS 2015, 2017, and 2018 datasets.

\subsection{Ablation Studies}
The training set of BraTS 2018 is randomly divided into two subsets to enable convenient evaluation. These two subsets are a training subset and a local validation subset, which consist of 260 and 25 MRI images respectively. Quantitative results of this local validation subset are presented in Table \ref{Table1}. Here, the one-model, two-model, and three-model cascades are denoted as MC1, MC2, and MC3, respectively. In the below, we conduct a series of experiments to prove the effectiveness of each component in the proposed approach.

\newcolumntype{I}{!{\vrule width 3pt}}
\newlength\savedwidth
\newcommand\whline{\noalign{\global\savedwidth\arrayrulewidth
		\global\arrayrulewidth 3pt}
	\hline
	\noalign{\global\arrayrulewidth\savedwidth}}
\newlength\savewidth
\newcommand\shline{\noalign{\global\savewidth\arrayrulewidth
		\global\arrayrulewidth 0.9pt}	
	\hline
	\noalign{\global\arrayrulewidth\savewidth}}	

\begin{table}[tp]
	\newcommand{\tabincell}[2]{\begin{tabular}{@{}#1@{}}#2\end{tabular}}	
	\fontsize{7}{14}\selectfont
	\centering
	\caption{Ablation Studies on the Local Validation Subset of BraTS 2018 }	
	\label{Table1}
	\begin{tabular}{c|c|c|c|c} 
		\shline
		\multirow{2}{*}{Method} & \multirow{2}{*}{Parameters} & \multicolumn{3}{c}{Dice (\%)} \\
		\cline{3-5}
		&  &Complete &Core &Enhancing  \\
		\hline 		
		MC1 &13.813 M  &90.41 &78.48 &72.91 \\
		\hline
		MC2 &27.626 M  &91.08 &79.11 &75.14 \\
		\hline
		MC3 &41.439 M  &91.08 &79.11 &79.53 \\
		\hline 
		\hline
		OM-Net &13.869 M  &91.10 &79.87 &\textbf{80.87} \\
		\hline		
		OM-Net$^0$ &13.869 M  &90.40 &79.41 &79.96 \\
		\hline
		OM-Net$^d$ &13.869 M  &91.11 &79.93 &80.26 \\	
		\hline
		\hline			
		OM-Net + SE & 13.870 M  &91.03 &80.20 &80.72 \\				
		\hline						
		OM-Net + CGA &13.814 M  &91.34 &82.15 &80.73 \\	
		\hline			
		OM-Net + CGA$^-$ &13.814 M  &91.06 &80.28 &80.78 \\		
		\hline						
		OM-Net + CGA$^t$ &13.814 M  &90.65 &80.27 &80.10 \\	
		\hline			
		OM-Net + CGA$^n$ &13.814 M  &89.75 &79.87 &76.00\\			
		\hline
		\hline
		OM-Net$^p$ &13.869 M  &91.28 &82.50 &80.84 \\	
		\hline	
		OM-Net + CGA$^p$ & 13.814 M  &\textbf{91.59} &\textbf{82.74} &80.73 \\	
		\shline
	\end{tabular}
\end{table}

\subsubsection{Effectiveness of Model Cascade Strategy}
From Table \ref{Table1}, we can observe that with the increase of model number in MC, Dice scores steadily improve. These results prove the contribution of each deep network in MC. Unfortunately, the number of parameters increases along with the number of models, leading to increased storage consumption and system complexity.

\subsubsection{Effectiveness of One-pass Multi-task Network}
We compare the performance of OM-Net with that of the MC strategy in Table \ref{Table1}. Despite having only one-third of the parameters of MC3, OM-Net consistently obtains better segmentation performance, especially for Dice scores on the tumor core and enhancing tumor. Moreover, we additionally train OM-Net$^0$ (a naive multi-task learning model without stepwise training or training data transfer) and OM-Net$^d$ (a multi-task learning model without stepwise training, but with training data transfer). It can be seen that OM-Net outperforms both OM-Net$^0$ and OM-Net$^d$; this demonstrates the effectiveness of the data transfer strategy and the curriculum learning-based training strategy.

\subsubsection{Effectiveness of Cross-task Guided Attention}
To compare the performance of the CGA module with that of the SE block, we further test the OM-Net + SE model where an SE block is inserted before each Classifier-$i$ $(1\leq i \leq 3)$ module of OM-Net in Fig. \ref{Fig.2}. Experimental results in Table \ref{Table1} show that OM-Net + CGA outperforms both OM-Net and OM-Net + SE. In particular, it outperforms OM-Net by as much as 2.28\% on Dice score for the tumor core region. Moreover, there is a slight performance drop for enhancing tumor of 0.14\%. However, on much larger data sets (see Table \ref{Table2}, \ref{Table3}, \ref{Table4}) where the experimental results are more stable, we observe that CGA consistently improves the Dice score on enhancing tumor. In comparison, there is no clear difference in performance between OM-Net and OM-Net + SE. This can be explained from two perspectives. First, the GAP operation in the SE block ignores category-specific statistics; second, recalibrating each channel with the same weight for all categories is suboptimal for segmentation. The proposed CGA module effectively handles the above two problems, and therefore achieves better performance than the SE block. In addition, the model size of OM-Net + CGA is smaller than both OM-Net and OM-Net + SE. We can thus safely attribute the performance gains to the CGA module rather than to the additional parameters.

\begin{figure*} 
	\centering
	\includegraphics[height=12.5cm,width=13cm]{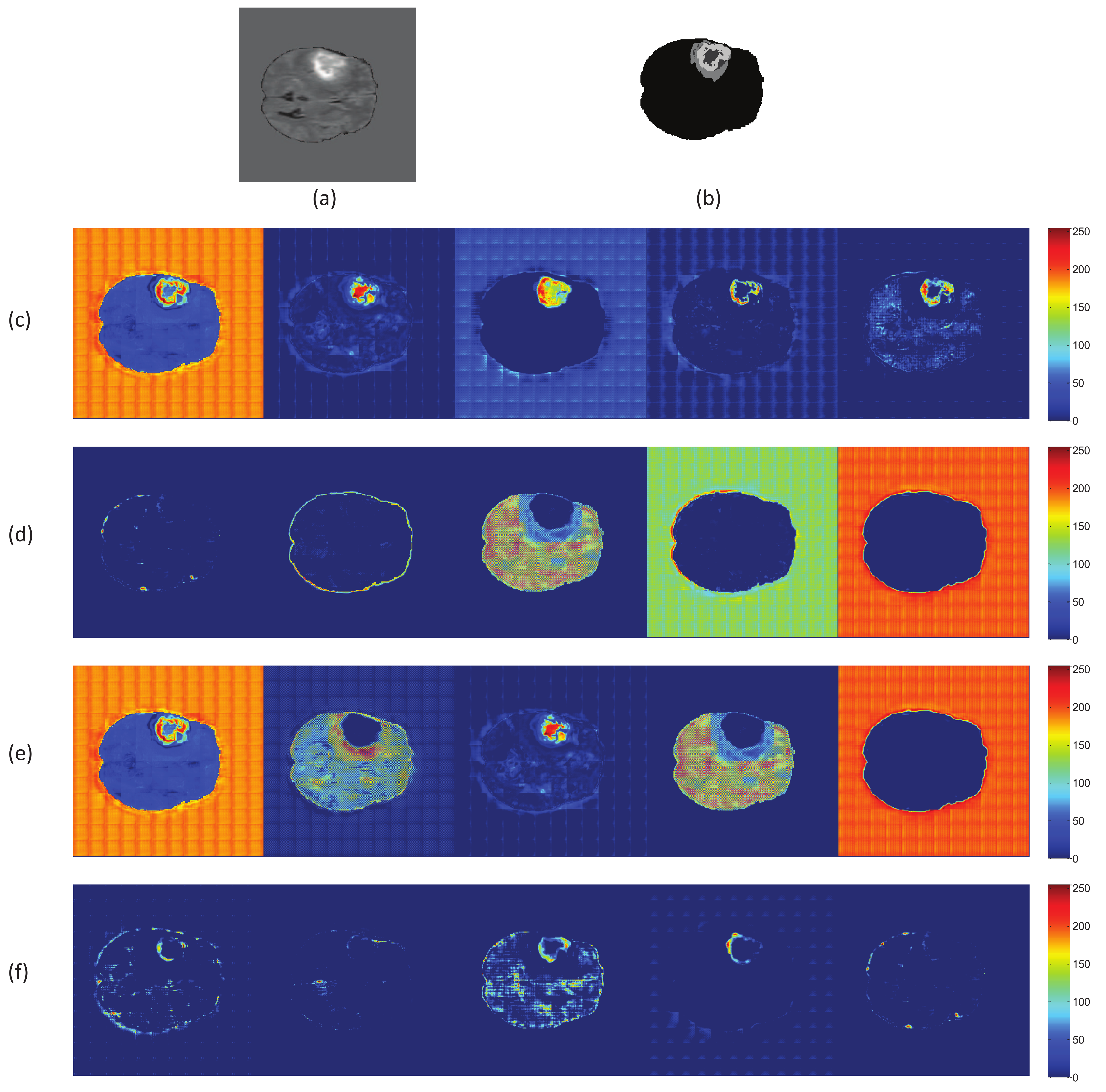}
	\caption{Visualization of the feature maps output from the shared backbone model in OM-Net + CGA. We present the feature maps of a complete 2D slice for intuitive and clear visualization. (a) The Flair modality on the $75^{th}$ slice of the sample \emph{BraTS18$\_$2013$\_$21$\_$1} in the BraTS 2018 training set. (b) Its corresponding ground truth. (c) Heat maps corresponding to the channels with the five largest values in $\mathbf{{m}_{t}}$. (d) Heat maps corresponding to the channels with the five smallest values in $\mathbf{{m}_{t}}$. (e) Heat maps corresponding to the channels with the five largest values in $\mathbf{{m}_{n}}$. (f) Heat maps corresponding to the channels with the five smallest values in $\mathbf{{m}_{n}}$. 					
	}
	\label{Fig.S1}
\end{figure*}

Next, we conduct additional experimental investigation into the CGA module in order to prove the validity of both the CSCI block and the CompSeg block.
\begin{itemize}
\item To justify the effectiveness of the CSCI block,
we visualize some feature maps produced by the shared backbone model, as shown in Fig. \ref{Fig.S1}. It should be noted that, in the interests of intuitive and clear visualization, we choose to visualize the feature maps of a complete 2D slice rather than those of a 3D patch. To achieve this, we stitch F-2 and O-1 of 3D patches in Fig. \ref{Fig.3} respectively to form the whole feature maps and probability maps. We then select the feature maps $\mathbf{F}$ and probability map $\mathbf{P}$ of a certain slice and calculate  $\mathbf{{m}_{t}}$ and $\mathbf{{m}_{n}}$ corresponding to this slice, according to Eqs. \ref{eq5} and \ref{eq5_1}, respectively. The channels corresponding to the five largest and five smallest values in $\mathbf{{m}_{t}}$ are presented in Fig. \ref{Fig.S1}(c) and Fig. \ref{Fig.S1}(d), respectively. Similarly, the channels with the five largest and five smallest values in $\mathbf{{m}_{n}}$ are presented in Fig. \ref{Fig.S1}(e) and Fig. \ref{Fig.S1}(f), respectively. As deconvolution layers are used in the model, there are inevitable checkboard artifacts; however, these do not affect the observations.

It is clear that the feature maps shown in Fig. \ref{Fig.S1}(c) do indeed have strong responses for the tumor region, which should be highlighted for the segmentation of the tumor region. In contrast, the feature maps in Fig. \ref{Fig.S1}(d) have weak responses for the tumor region, but strong responses for the non-tumor region; therefore, they will be suppressed in CGA for the segmentation of the tumor region.  Similarly, consistent observations can be found in Fig. \ref{Fig.S1}(e) and Fig. \ref{Fig.S1}(f). The above analysis proves the validity of the CSCI block in generating the category-specific channel dependence.

\item Furthermore, we also justify the effectiveness of the CompSeg block, where $\mathbf{P_t}$ and $\mathbf{P_n}$ are used a second time. We additionally train a model without using $\mathbf{P_t}$ and $\mathbf{P_n}$, which simply performs an element-wise addition between $\mathbf{S_t}$ and $\mathbf{S_n}$ in Fig. \ref{Fig.4}(b), denoted as OM-Net + CGA$^-$.
Experimental results in Table \ref{Table1} reveal that OM-Net + CGA significantly outperforms OM-Net + CGA$^-$; this is because OM-Net + CGA employs soft spatial masks ($\mathbf{P_t}$ and $\mathbf{P_n}$) to fuse two complementary prediction results. This performance comparison proves the validity of $\mathbf{P_t}$ and $\mathbf{P_n}$ used in the CompSeg block. In addition, we also test another two models: OM-Net + CGA$^{t}$ is a model whose CompSeg block only includes the upper branch in Fig. \ref{Fig.4}(b), while OM-Net + CGA$^{n}$ is a model whose CompSeg block only incorporates the lower branch in Fig. \ref{Fig.4}(b). The experimental results are reported in Table \ref{Table1}. From the table, it can be seen that OM-Net + CGA outperforms both OM-Net + CGA$^{t}$ and OM-Net + CGA$^{n}$. Accordingly, we can conclude that the two pathways in the CompSeg block can make full use of the complementary information, which is beneficial for the segmentation task.
\end{itemize}

\subsubsection{Effectiveness of Post-processing}
To justify the effectiveness of the proposed post-processing operation, we apply it to refining the segmentation results of both OM-Net and OM-Net + CGA, denoted as OM-Net$^p$ and OM-Net + CGA$^p$ respectively in Table \ref{Table1}. First, compared with OM-Net, it is shown that OM-Net$^p$ can slightly improve the Dice score for the complete tumor due to false positives having been removed in the first post-processing step; meanwhile, it significantly improves the Dice score of tumor core by 2.6\% because of the second post-processing step. Moreover, the performance comparison between OM-Net + CGA and OM-Net + CGA$^p$ shows that post-processing operation consistently brings about performance improvement for the complete tumor and tumor core regions.

\begin{figure*}
	\setlength{\belowcaptionskip}{-60.cm}
	\centering 
	\includegraphics[height=10cm,width=9.5cm]{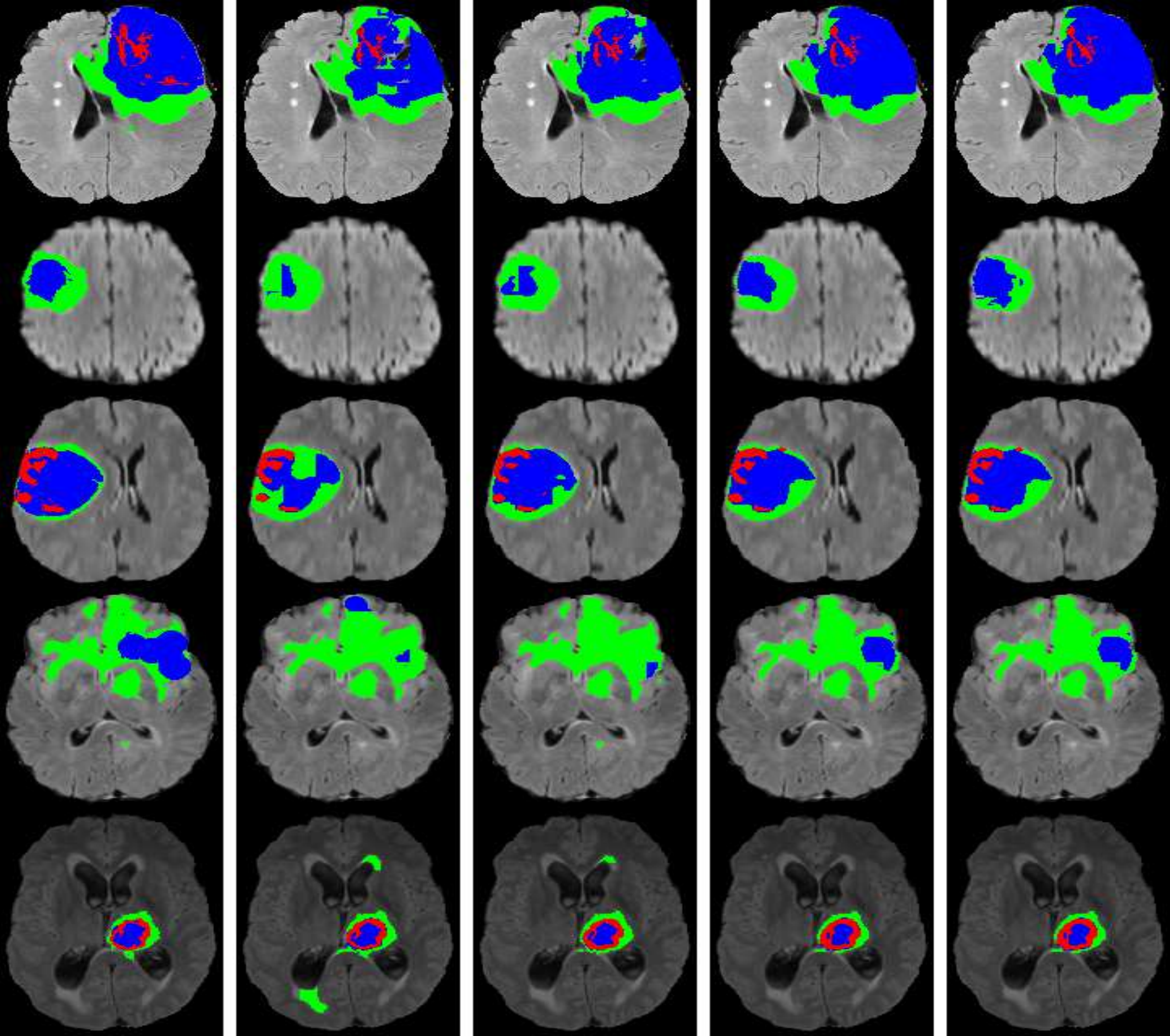}
	\caption{Example segmentation results on the local validation subset of BraTS 2018. From left to right: Ground truth, MC3, OM-Net, OM-Net + CGA, and OM-Net + CGA$^{p}$ results overlaid on FLAIR image; edema (green), necrosis and non-enhancing (blue), and enhancing(red).
	}
	\label{Fig.S2}
	
\end{figure*}

In conclusion, the above experimental results justify the effectiveness of the proposed techniques. Qualitative comparisons between MC3, OM-Net, OM-Net + CGA, and OM-Net + CGA$^p$ are also provided in Fig. \ref{Fig.S2}. It is clear that the proposed methods steadily improve the quality of brain tumor segmentation. This is consistent with the quantitative comparisons in Table \ref{Table1}.

\subsection{Performance Comparison on BraTS 2015 Testing Set}
In this experiment, the performance of MC3, OM-Net, and OM-Net + CGA is evaluated on the testing set of the BraTS 2015 dataset. Each of these models is trained using the entire training set of the dataset. Experimental results are tabulated in Table \ref{Table2}. From this table, we can make the following observations.

First, we compare the segmentation performance of MC3, OM-Net, OM-Net + CGA, and OM-Net + CGA$^p$. It is evident that OM-Net has clear advantages over MC3, with 1\% higher Dice scores on both tumor core and enhancing tumor. OM-Net + CGA further promotes the Dice score of OM-Net by 1\% on enhancing tumor. Moreover, following refinement by the proposed post-processing method, the Dice scores of OM-Net + CGA improve significantly by 1\% and 4\% on complete tumor and tumor core, respectively. These results are consistent with comparisons on the local validation subset of BraTS 2018.

Second, we compare the performance of OM-Net + CGA$^p$ with state-of-the-art methods. Our results exhibit clear advantages over those obtained by the comparison methods \cite{isensee2017brain,chen2018focus, kamnitsas2017efficient,zhao2018deep}. In particular, our results outperform the popular DeepMedic model \cite{kamnitsas2017efficient} by 2\%, 8\%, and 2\% in terms of Dice scores on complete tumor, tumor core, and enhancing tumor, respectively. Furthermore, OM-Net + CGA$^p$ also outperforms the method in \cite{zhao2018deep} which adopts more pre-processing and post-processing operations. At the time of this submission, OM-Net + CGA$^p$ ranks first on the online leaderboard of BraTS 2015, demonstrating the effectiveness of the proposed methods.

\subsection{Performance Comparison on BraTS 2017 Validation Set}
Since access to the testing set of BraTS 2017 was closed after the challenge, we evaluate the proposed methods on the online validation set and compare them with other participants in Table \ref{Table3}.

First, we train the MC3, OM-Net, OM-Net + SE, and OM-Net + CGA models using the training subset of 260 MRI images. The following observations can be made. OM-Net outperforms MC3, especially for the Dice scores on enhancing tumor. The SE block cannot improve the performance of OM-Net in terms of Dice scores; in fact, it reduces the Dice score of OM-Net by 1.18\% on enhancing tumor. In comparison, OM-Net + CGA outperforms both OM-Net and OM-Net + SE by a significant margin. Specifically, its Dice scores are 1.88\% and 2.09\% higher than those of OM-Net on tumor core and enhancing tumor respectively; this again proves the effectiveness of the CGA module. In addition, OM-Net + CGA$^p$ considerably improves the Dice score on tumor core, demonstrating the effectiveness of the proposed post-processing method.

Second, to further boost the performance of OM-Net + CGA, we divide the training data into ten folds and obtain an ensemble system using the following scheme: we train one model using nine folds of training data and pick the best snapshot on the rest one fold. We repeat this process and obtain 10 models as an ensemble (denoted as OM-Net + CGA$^\star$). A similar strategy was used in \cite{isensee2018no}. Table \ref{Table3} shows that OM-Net + CGA$^\star$ consistently obtains higher Dice scores than OM-Net + CGA. We further apply the proposed post-processing method to OM-Net + CGA$^\star$, denoted as OM-Net + CGA${^\star}^p$. It can clearly be seen that the proposed post-processing method improves the Dice score of OM-Net + CGA${^\star}$  by as much as 1.48\% for tumor core.

Third, we present comparisons between OM-Net + CGA${^\star}^p$ and some state-of-the-art methods on the online validation leaderboard, which comprises more than 60 entries. It is clear that OM-Net + CGA${^\star}^p$ outperforms all other methods in terms of Dice scores. It is further worth noting that other top entries, such as Kamnitsas \emph{et al.} \cite{kamnitsas2017ensembles}, also combined multiple models to boost performance; moreover, Wang \emph{et al.} \cite{wang2017automatic} integrated nine single-view models from three orthogonal views to achieve excellent performance. The above comparisons demonstrate the superiority of our proposed methods.

\begin{table*}
	\newcommand{\tabincell}[2]{\begin{tabular}{@{}#1@{}}#2\end{tabular}}
	\fontsize{7}{14}\selectfont
	\centering
	\caption{Performance on BraTS 2015 Testing Set  (\%) }
	
	\label{Table2}
	\begin{tabular}{c|p{1.05cm}<{\centering}|p{0.75cm}<{\centering}|p{1.15cm}<{\centering}|p{1.05cm}<{\centering}|p{0.75cm}<{\centering}|p{1.15cm}<{\centering}|p{1.05cm}<{\centering}|p{0.75cm}<{\centering}|p{1.15cm}<{\centering}}  
		\shline
		\multirow{2}{*}{Method} & \multicolumn{3}{c|}{Dice} &\multicolumn{3}{c|}{Positive Predictive Value} & \multicolumn{3}{c}{Sensitivity}\\
		\cline{2-10}
		&Complete &Core &Enhancing &Complete &Core &Enhancing &Complete &Core &Enhancing \\
		\hline
		MC3 &86 &70 &63 &86 &82 &60 &88 &67 &72\\
		\hline
		OM-Net &86 &71 &64 &86 &83 &61 &88 &68 &72\\
		\hline		
		OM-Net + CGA &86 &71 &65 &87 &84 &63 &88 &67 &70\\	
		\hline
		OM-Net + CGA$^p$ &\textbf{87} &\textbf{75} &\textbf{65} &89 &85 &63 &88 &73 &70\\			
		\hline 
		\hline
		Isensee \emph{et al.} \cite{isensee2017brain} &85 &74 &64 &83 &80 &63 &91 &73 &72\\
		\hline
		Chen \emph{et al.} \cite{chen2018focus} &85 &72 &61 &86 &83 &66 &86 &68 &63\\
		\hline	
		Zhao \emph{et al.} \cite{zhao2018deep} &84 &73 &62 &89 &76 &63 &82 &76 &67\\
		\hline	
		Kamnitsas \emph{et al.} \cite{kamnitsas2017efficient} &85 &67 &63 &85 &86 &63 &88 &60 &67\\
		\shline
	\end{tabular}
	
\end{table*}

\begin{table}
	\newcommand{\tabincell}[2]{\begin{tabular}{@{}#1@{}}#2\end{tabular}}	
	\fontsize{6.5}{14}\selectfont
	\centering
	\caption{Mean Values of Dice and Hausdorff95 Metrics on BraTS 2017 Validation Set }	
	\label{Table3}
	\begin{threeparttable}
		\begin{tabular}{p{2cm}<{\centering}|p{0.65cm}<{\centering}|p{0.65cm}<{\centering}|p{0.65cm}<{\centering}|p{0.65cm}<{\centering}|p{0.65cm}<{\centering}|p{0.65cm}<{\centering}}
			\shline
			\multirow{2}{*}{Method} & \multicolumn{3}{c|}{Dice} & \multicolumn{3}{c}{Hausdorff95 (mm)}\\ 
			\cline{2-7}
			&Enh. &Whole &Core &Enh. &Whole &Core   \\
			\hline  
			MC3 &0.7424 &0.8991 &0.7937 &4.9901 &4.6085 &8.5537  \\			
			\hline  
			OM-Net &0.7534 &0.9007 &0.7934 &3.6547 &7.2524 &8.4676  \\
			\hline
			OM-Net + SE  &0.7416 &0.8997 &0.7938 &3.5115 &6.2859 &7.0154 \\
			\hline
			OM-Net + CGA  &0.7743 &0.8988 &0.8122 &3.8820 &4.8380 &6.7953  \\
			\hline
			OM-Net + CGA$^p$  &0.7743 &0.9016 &0.8320 &3.8820 &4.6663 &6.7312  \\			
			\hline
			OM-Net + CGA\tnote{$\star$}   &0.7852 &0.9065 &0.8274 &3.2991 &4.4886 &6.9896  \\
			\hline
			OM-Net + CGA${^\star}^p$ &\textbf{0.7852} &\textbf{0.9071} &\textbf{0.8422} &3.2991 &4.3815 &7.5614  \\			
			\hline
			\hline		
			Wang \emph{et al.} \cite{wang2017automatic}  &0.7859 &0.9050 &0.8378 &3.2821 &3.8901   &6.4790\\ 
			\hline
			MIC$\_$DKFZ &0.7756 &0.9027 &0.8194 &3.1626 &6.7673 &8.6419 \\
			\hline
			inpm &0.7723 &0.8998 &0.8085 &4.7852 &9.0029 &7.2359 \\
			\hline
			xfeng &0.7511 &0.8922 &0.7991 &4.7547 &16.3018 &8.6847  \\
			\hline
			Kamnitsas \emph{et al.} \cite{kamnitsas2017ensembles}  &0.738 &0.901 &0.797 &4.50 &4.23  &6.56\\ 	        	
			\shline			
			
		\end{tabular}
		
	\end{threeparttable}
\end{table}

\subsection{Performance Comparison on BraTS 2018 Dataset}
We also make additional comparisons on the BraTS 2018 dataset. As BraTS 2018 and 2017 share the same training dataset, we directly evaluate the same models in the previous experiments on the validation set of BraTS 2018. The BraTS 2018 Challenge is intensely competitive, with more than 100 entries displayed on the online validation leaderboard. Therefore, we only present comparisons between our methods and the top entries in Table \ref{Table4}. Our observations are as follows.

\begin{table}
	\newcommand{\tabincell}[2]{\begin{tabular}{@{}#1@{}}#2\end{tabular}}	
	\fontsize{6.5}{14}\selectfont
	\centering
	\caption{Mean Values of Dice and Hausdorff95 Metrics on BraTS 2018 Validation Set }
	\label{Table4}
	\begin{threeparttable}
		\begin{tabular}{p{1.8cm}<{\centering}|p{0.65cm}<{\centering}|p{0.65cm}<{\centering}|p{0.65cm}<{\centering}|p{0.65cm}<{\centering}|p{0.65cm}<{\centering}|p{0.65cm}<{\centering}}%
			\shline
			\multirow{2}{*}{Method} & \multicolumn{3}{c|}{Dice} & \multicolumn{3}{c}{Hausdorff95 (mm)}\\ 
			\cline{2-7}
			&Enh. &Whole &Core &Enh. &Whole &Core   \\
			\hline  
			MC3 &0.7732 &0.9015 &0.8233 &4.1624 &4.7198 &7.6082  \\				
			\hline  
			OM-Net &0.7882 &0.9034 &0.8273 &3.1003 &6.5218 &7.1974  \\
			\hline
			OM-Net + SE  &0.7791 &0.9034 &0.8259 &2.9950 &5.7685 &6.4289 \\
			\hline
			OM-Net + CGA  &0.8027 &0.9033 &0.8399 &3.4437 &4.7609 &6.4339  \\
			\hline
			OM-Net + CGA$^p$  &0.8027 &0.9052 &0.8536 &3.4437 &4.6236 &6.3892  \\				
			\hline
			OM-Net + CGA\tnote{$\star$}   &0.8112 &0.9074 &0.8461 &2.8697 &4.9105 &6.6243  \\
			\hline
			OM-Net + CGA${^\star}^p$ &0.8111 &0.9078 &0.8575 &2.8810 &4.8840 &6.9322  \\				
			\hline	
			\hline				
			Myronenko  \cite{myronenko20183d} &0.8233 &0.9100 &0.8668 &3.9257 &4.5160 &6.8545\\ 
			\hline			     		
			SHealth &0.8154 &0.9120 &0.8565 &4.0461 &4.2362 &7.2181 \\
			\hline
			Isensee \emph{et al.} \cite{isensee2018no} 
			$^{\dagger}$
			&0.8048 &0.9072 &0.8514 &2.81 &5.23 &7.23 \\	 		 
			\hline				
			MedAI &0.8053 &0.9104 &0.8545 &3.6695 &4.1369 &5.9821  \\
			\hline
			BIGS2 &0.8054 &0.9104 &0.8506 &2.7543 &4.8444 &7.4548  \\
			\hline			
			SCAN &0.7925 &0.9008 &0.8474 &3.6035 &4.0626 &4.9885  \\			
			\shline				
		\end{tabular}
		\begin{tablenotes}
			\item[$^{\dagger}$]To facilitate fair comparison, we report the performance of \cite{isensee2018no} without private training data.	
		\end{tablenotes}
	\end{threeparttable}
\end{table}

First, comparison results between MC3, OM-Net, OM-Net + SE, and OM-Net + CGA are consistent with those in Table \ref{Table3}. 
We can see that OM-Net achieves better performance than MC3 on enhancing tumor with a visible margin. Moreover, despite having only a single model and without post-processing, OM-Net is able to outperform more than 70\% of entries on the leaderboard. In addition, OM-Net + CGA outperforms OM-Net by 1.26\% and 1.45\% in terms of Dice scores on tumor core and enhancing tumor, respectively. By comparison, SE cannot improve the performance of OM-Net in terms of Dice scores.

Second, by implementing the model ensemble and the post-processing operation, OM-Net + CGA${^\star}^p$ obtains higher Dice scores as expected, achieving very competitive performance on the leaderboard. It is worth noting that the model ensemble strategy was also applied in \cite{kamnitsas2017ensembles,myronenko20183d,isensee2018no}. The approach described in \cite{myronenko20183d} also decomposes the multi-class brain tumor segmentation into three tasks. It achieves top performance by training 10 models as an ensemble, the inputs of which are large patches of size $160 \times 192 \times 128$ voxels. Large input patches lead to considerable memory consumption; therefore, 32GB GPUs are employed in \cite{myronenko20183d} to train this model. In comparison, OM-Net utilizes small patches of size $32 \times 32 \times 16$ voxels, making it memory-efficient and capable of being trained or deployed on low-cost GPU devices. We can thus conclude that OM-Net is very competitive and has its own advantages.

More impressively, benefitting from the techniques proposed in this paper, we obtained the joint third position among 64 teams on the testing set of the BraTS 2018 Challenge\footnote[2]{\url{https://www.med.upenn.edu/sbia/brats2018/rankings.html}}. More detailed results of this challenge are introduced in \cite{bakas2018identifying}. In conclusion, the effectiveness of the proposed methods has been demonstrated through comparisons on the above three datasets.

\section{Conclusion}
In this paper, we propose a novel model, OM-Net, for brain tumor segmentation that is tailored to handle the class imbalance problem. Unlike the popular MC framework, OM-Net requires only one-pass computation to perform coarse-to-fine segmentation. OM-Net is superior to MC because it not only significantly reduces the model size and system complexity, but also thoroughly exploits the correlation between the tasks through sharing parameters, training data, and even prediction results. In particular, we propose a CGA module that makes use of cross-task guidance information to learn category-specific channel attention, enabling it to significantly outperform the popular SE block. In addition, we introduce a novel and effective post-processing method for use in refining the segmentation results in order to achieve better accuracy. Extensive experiments were conducted on three popular datasets; the results of these experiments prove the effectiveness of the proposed OM-Net model, and further demonstrate that OM-Net has clear advantages over existing state-of-the-art methods for brain tumor segmentation.

\end{document}